# Impact Invariant Trajectory Optimization of 5-Link Biped Robot Using Hybrid Optimization

Aref Amiri, Hasan Salarieh[1]

*Department of Mechanical Engineering, Sharif University of Technology, Tehran, Iran*

**Abstract**

Bipedal robots have received much attention because of the variety of motion maneuvers that they can produce, and the many applications they have in various areas including rehabilitation. One of these motion maneuvers is walking. In this study, we presented a framework for the trajectory optimization of a 5-link (planar) Biped Robot using hybrid optimization. The walking is modeled with two phases of single-stance (support) phase and the collision phase. The dynamic equations of the robot in each phase are extracted by the Lagrange method. It is assumed that the robot heel strike to the ground is full plastic. The gait is optimized with a method called hybrid optimization. The objective function of this problem is considered to be the integral of torque-squared along the trajectory, and also various constraints such as zero dynamics are satisfied without any approximation. Furthermore, in a new framework, there is presented a constraint called impact invariance, which ensures the periodicity of the time-varying trajectories. On the other hand, other constraints provide better and more human-like movement..

## 1. Introduction

The mechanism of movement and transfer of objects has always been one of the most important and active areas of human research. Due to the limitations of moving with a wheel, replacing it with feet is an attractive but difficult option, so this field is a hot topic in today's robotic world. With the advancement of robotics science and the usefulness of this issue, a lot of research has been done on the design, optimization, and control of legged robots [1-6]. As the science of bipedal robots has advanced in recent years, there have been significant efforts to improve the performance of these robots in important maneuvers, such as walking and running, but research is still ongoing to find ideal answers [7,8]. Designing reference trajectories for human walking cycles is very important. Several techniques have been adopted to define reference trajectories. So far, many researchers have studied low-energy (or low input torques) paths for bipedal robots [7,9]. We are looking for a periodic path that meets a specific goal in terms of speed and minimizes the torque required to produce the gate. In general, this open and non-trivial problem is solved by finding numerical answers. Various parameters can be considered to optimize the problem, for example, torques, Cartesian coordinate or joint coordinates constraints can be used[10-12]. Many authors have used polynomial functions for Cartesian coordinates of swing leg's foot, hip, and trunk angle [13,14]. Polynomial functions are used for the coordinates of the joints to limit the number of optimization parameters [15]. The optimal path for each coordinate of joints is usually written in the form of polynomials with unknown coefficients. The coefficients should be obtained through the optimization process [15]. For all bipedal robots, it is important to define optimal periodic motions despite the fact that the number of actuators is less than the degree of freedom of the system, and also zero dynamics problem there exists which should be satisfied during optimization.

In this paper, a new method is presented to produce a periodic path for the walking of bipedal robots which satisfies the impact invariance constraint. Also, in order to achieve the feasible trajectory, the zero dynamics constraint is satisfied without any approximation. In addition, by considering some other kinematic and dynamic constraints, and using the hybrid optimization method, an optimal reference trajectory for human-like walking of bipedal robots has been presented.

[1]P.O.B. 11155-9567, Tehran, Iran
 salarieh@sharif.edu

Section 2 presents the dynamics and kinematics of a model of the biped robot. Section 3 is devoted to the formulation of the optimization variables. The constraints are defined in section 4 and the optimization method is also described in section 5. Finally, in sections 5 and 6, the results and conclusion are given.

## 2. Dynamics and kinematics

Bipedal robots have different dynamics depending on their movement maneuvers. For example, a running robot with 5 links and without an ankle actuator in the flight phase has 7 degrees of freedom and only 4 actuators, so the system is 3 degrees under-actuated. Here a bipedal walking robot will be examined. We assume that the robot is completely on the ground and does not slip while walking.

On the other hand, during the single support phase, the other leg rises from the ground when the swing leg hits the ground. During the single support phase, the model has 5 degrees of freedom and needs at least 5 generalized coordinates to identify the system. On the other hand, the robot has only 4 actuation, so the system has a degree of under-actuation. In under-actuated systems, some parts of the dynamics are not affected by the actuator called the zero dynamics. Here, zero dynamics is affected only by the earth's gravity. The robot's model can be modeled with absolute or relative angles, if relative angles are used, zero dynamics can be easily separated from the main total dynamics. Figure 1 shows the absolute and relative coordinates of a 5-link robot with point feet.

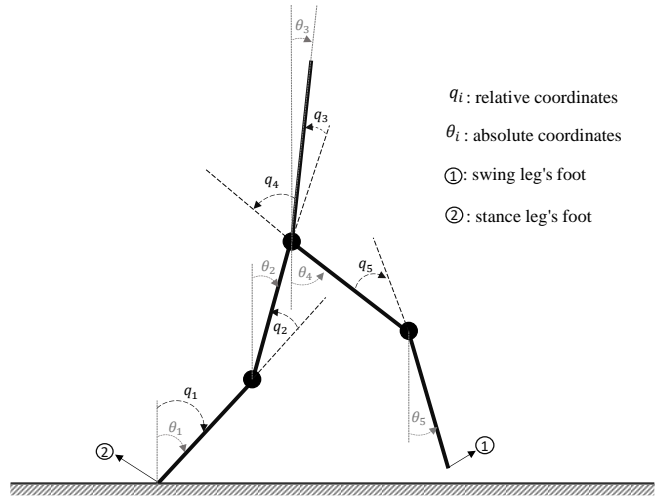

*Figure 1 Relative and absolute angles*

The general hybrid walking gait model is obtained by combining the single support phase model and the impact model:

$$\Sigma: \begin{cases} \dot{x} = f(x) + g(x)u & x^- \notin \Gamma \\ x^+ = \tilde{\Delta}(x^-) & x^- \in \Gamma \end{cases} \quad (1)$$

where $\tilde{\Delta}$ is a mapping that transforms the states just before the contact to the states just after the contact. $x := (q^T, \dot{q}^T)^T$ is the state vector that contains $q := (q_1, q_2, \dots, q_n)^T$ which is the vector of joint coordinates and $\dot{q} := (\dot{q}_1, \dot{q}_2, \dots, \dot{q}_n)^T$ is the vector of angular velocities, and $x^+$ denotes the state vector just after the impact and $x^-$ shows just before this event.

The switching set is shown as,

$$\Gamma = \{(q,\dot{q}) \in x \mid P^v(q) = 0, P^h(q) > 0\} \tag{2}$$

$P^v(q)$ and $P^h(q)$ indicate the vertical and horizontal position of the swing leg, respectively. Now if we model the single support phase alone, we have:

$$M(q)\ddot{q} + c(q,\dot{q})\dot{q} + G(q) = (0, U^T)^T \tag{3}$$

where $M(q) \in \Re^{n \times n}$ ($n = 5$) is the inertia matrix, $c(q,\dot{q}) \in \Re^{n \times n}$ is the Coriolis matrix, and $G(q) \in \Re^n$ is the gravity vector. As shown in Figure 2, the robot does not have any actuators (torques) on the feet, i.e. the robot has not the ankle joint actuator, so the robot is under-actuated which adds a zero dynamic constraint to the problem as mentioned in [16]. The vector $U \in \Re^{n-1}$ is as follows:

$$U = [\tau_1, \tau_2, \tau_3, \tau_4]^\tau \tag{4}$$

which represents 4 actuators (torques) on the robot. 2 actuators (torques) on the pelvis (hip) and 2 on the knee of each leg. By separating the equations of (3) the first equation which produces the zero dynamics is written as:

$$\sum_{j=1}^{5} \left(M_{1,j}\ddot{q}_j + c_{1,j}\dot{q}_j\right) + G_1 = 0 \tag{5}$$

which is called zero-hybrid dynamics and:

$$\sum_{j=1}^{5} \left(M_{i,j}\ddot{q}_j + c_{i,j}\dot{q}_j\right) + G_i = \tau_{i-1} \tag{6}$$

are other rows of equation (3) (i = 2,…, 5).

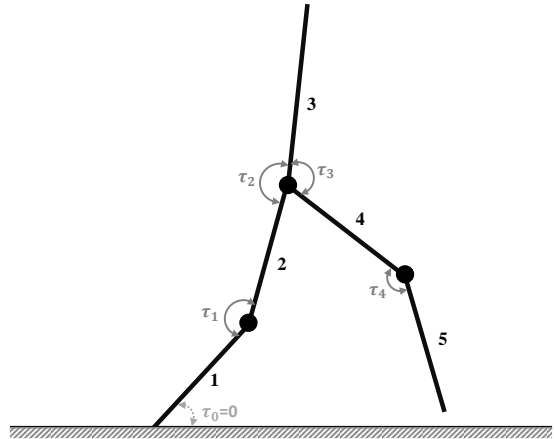

*Figure 2 Robot configuration and control torques*

The trunk angle is assumed independent from other links with a separate actuator, In other words, one actuator is responsible for moving the trunk. So if we temporarily separate the trunk from the other components, we are faced with 4 degrees of freedom system. By determining the swing leg's foot (link number 5 in figure 2), the system still has 2 degrees of freedom, so the inverse kinematics has infinite answers. Therefore, By determining the position of the hip, 2 more degrees of freedom are determined from the system, in this case, the inverse kinematic robot has 4 answers. Among these 4 answers, the only acceptable answers are the one that satisfies the condition of not breaking the knee. It is important to note that in order to find a suitable periodic answer, we assume that the initial configuration is the same as the final one.

## 3. Optimization variables

One convenient way is to select the angles of each link based on a polynomial function of time with a series of unknown coefficients. This choice enables us to have a smooth function with time. Here it is assumed that each angle is a polynomial function of degree 4. It should be noted that the initial and final configuration of the system in each step affects determining two parameters of the polynomial coefficients, the impact invariance constraint is also effective on another coefficient. Therefore, in order to have at least 2 optimization parameters for each angle, we consider a fourth-order polynomial function with unknown coefficients for the trajectories of each angle.

$$q_k(t) = \sum_{i=0}^{n=4} \alpha_{k,i} t^i \quad (k = 1, \ldots, 5) \tag{7}$$

## 4. Definition of constraints

These constraints are to find the right trajectory to walk. It makes the shapes of the joint trajectories, the links orientations, and the required torques for walking be within a reasonable range.

The constraints are defined as follows:

1) Constraints on the initial and final configuration: Initial and final configurations of the robot must be specified. Since the robot moves in a periodic pattern, its initial and final configuration must coincide.

$$q_{(@t=0)} = q_{initial}, \quad q_{(@t=T)} = q_{final}, \tag{8}$$

2) Knee movement constraints: In order to have human-like movement, the robot's knees should not be opened and closed excessively ( $m_1$ and $m_2$ are two pre-especified upper bounds in Eq. (9)).

$$m_1 \geq q_{2(t)} \geq 0, \quad m_2 \geq q_{4(t)} \geq 0, \tag{9}$$

3) Swing leg's foot constraint: The swing leg's foot should not collide with the ground except at the beginning and end of the phase.

$$p_{(0)}^v = p_{(T)}^v = 0 \quad p_{(t)}^v > 0 \text{ for } 0 < t < T \tag{10}$$

4) Limitation of torques: In order to the physical limitations of the motors, the actuator torques have a certain limit.

$$|\tau_{i-1(t)}| \leq \tau_{max} \quad i = 2, \ldots, 5 \tag{11}$$

5) Limitation of angular velocities: In order to the physical limitations of the motors, the actuator velocities have a certain limit.

$$|\dot{q}_{i(t)}| \leq \dot{q}_{max} \quad i = 1, \ldots, 5 \tag{12}$$

6) Limitation of friction coefficient: The reaction of the heels, which is the result of the acceleration of the various members of the robot, must observe a certain ratio. This ratio should be less than the coefficient of friction between the heels and the ground.

$$-\mu \leq \left|\frac{F_x}{F_y}\right| \leq \mu \tag{13}$$

In the above equation, $\mu$ is the coefficient of friction, and $F_x$ and $F_y$ are sequentially the horizontal and vertical ground reactions.

7) Zero dynamic constraint: the satisfaction of this constraint is important in two ways. First, if this constraint is not satisfied, the problem of optimizing the input torques is practically ambiguous, because these torques are not really applicable to the problem. Although it may lead to a feasible kinematic equation (kinematically possible), it is not feasible in terms of control, or in other words, it is not dynamically possible.

8) Impact invariance constraint: this constraint means that in order to produce a periodic motion, in addition to the configuration, the initial velocities at the beginning point of each cycle should be exactly the same as its previous cycle. Since the velocities after the collision are dependent on the velocities before the collision, by satisfying this constraint, the velocities before the collision are adjusted in such a way as to guarantee the periodicity of the motion. Through the following formulae, this purpose is achieved. At first, the impact mapping formula is written as,

$$\dot{q}^+ = \tilde{\Delta}(q^-)\dot{q}^- \tag{14}$$

$\tilde{\Delta}(q^-) \in \Re^{5 \times 5}$ is the impact mapping which maps the angular rates of the leg before contact to the angular rates of that leg after contact. The inverse of $\tilde{\Delta}$ is denoted by,

$$\tilde{\eta}(q^-) = \left(\tilde{\Delta}(q^-)\right)^{-1} \tag{15}$$

So $\dot{q}^-$ can be found as :

$$\dot{q}^- = \tilde{\eta}(q^-)\dot{q}^+ \tag{16}$$

The mathematical formulation of this mapping is obtained from the governing differential equations of the system. After the swing leg's foot hits the ground, the positions do not change but the angular velocities change, which can be achieved as following (see [17] for more information),

$$\Delta \dot{q} = M^{-1} \cdot J^T \cdot (J \cdot M^{-1} \cdot J^T)^{-1} \cdot \Delta v_e \tag{17}$$

where $v$ is the velocity vector of the end of the swing leg and $M \in \Re^{n \times n}$ is the inertia matrix as mentioned in (3), the matrix $J \in \Re^{m \times n}$ ($m = 2$ for planar motions) is also obtained as:

$$J = \frac{\partial p_e}{\partial q} \tag{18}$$

$p_e$ is the position of the end of the swing leg. Assuming that the swing leg sticks to the ground after impact, the velocity of the swing leg's foot after impact is zero, so

$$\dot{q}^+ = \dot{q}^- + M^{-1} \cdot J^T \cdot (J \cdot M^{-1} \cdot J^T)^{-1} \cdot (-v_e) \tag{19}$$

We know that due to the placement of a leg on the ground, we can write:

$$v_e = \alpha_{(q)} \dot{q} \tag{20}$$

where $\alpha_{(q)}$ is:

$$\alpha_{(q)} = \frac{\partial v_e}{\partial \dot{q}} \tag{21}$$

Finally, by placing (20) into (19) and separating $\dot{q}^-$, the pre-impact angular velocity is obtained as follows:

$$\dot{q}^- = \left(I + M^{-1} \cdot J^T \cdot (J \cdot M^{-1} \cdot J^T)^{-1} - \alpha_{(q)}\right)^{-1} \dot{q}^+ \tag{22}$$

where $I \in \Re^{n \times n}$ is the identity matrix. In the above relation, both velocity vectors are written in the same coordinate system, which requires a coordinate conversion, because the coordinate changes after the collision due to the change in the role of the legs. For this purpose, consider the following mapping that converts the relative angles and angular velocities to absolute ones:

$$^{rel}\kappa = H \ ^{abs}\kappa \tag{23}$$

where $\kappa \in \Re^n$ can be the angles vector, the angular velocities vector or the angular accelerations vector. Superscripts $^{rel}\square$ and $^{abs}\square$ represent relative and absolute coordinates in which the vectors are defined, and also $H \in \Re^{n \times n}$ is a square matrix. On the other hand, we have a mapping that converts old and new coordinates to each other. This mapping can just be defined for an absolute angular coordinate. If we define the absolute coordinates in this way, we have:

$$_2\psi = \Gamma_1 \psi \tag{24}$$

where indices 1 and 2 indicate the coordinate system before and after the impact, $\psi \in \mathfrak{R}^{n \times n}$ can be velocity vector or angular acceleration vector, and $\Gamma \in \mathfrak{R}^{n \times n}$ is the mapping matrix. Finally, with the above transformations, the coordinate systems can be connected suitably as:

$$_2\dot{q}^+ = H\Gamma H^{-1}\,_1\dot{q}^+ \tag{25}$$

So the invariancy of the impact during walking is written as it follows,

$$\dot{q}^- = \left((I + M^{-1} \cdot J^T \cdot (J \cdot M^{-1} \cdot J^T)^{-1} - \alpha_{(q)}\right)^{-1} H\Gamma H^{-1}\dot{q}^+ \tag{26}$$

As a result, according to Equation (26), the impact invariance constraint is obtained. In this way, by satisfying this equality constraint, the velocity after impact will be similar to the initial velocity in the previous cycle.

## 5. Optimization

According to figure 3, optimization is performed using a hybrid method. This means that first, with the penalty method, the constrained problem becomes unconstrained. Then, using the genetic algorithm, the first level of optimization is applied. Finally, in the second level, the outputs of the first level are used as the input of a gradient-based method and the problem is solved. The objective function is the Euclidean norm of input torques:

$$J(\alpha) = \int_0^{T(\zeta^-)} \|U_\alpha(t)\|_2^2 dt = \int_0^{T(\zeta^-)} \langle \tau, \tau \rangle dt \tag{27}$$

where $T(\zeta^-)$ corresponds to the step duration, $U_\alpha(t)$ is the resulting torque obtained from (3) along the periodic solution of the hybrid zero dynamics. To solve the problem more easily and accurately, we tried to satisfy configuration constraints in the problem itself. Therefore, 2 coefficients of each coordinate and a total of 10 parameters of equation (7) are determined by the configuration constraints.

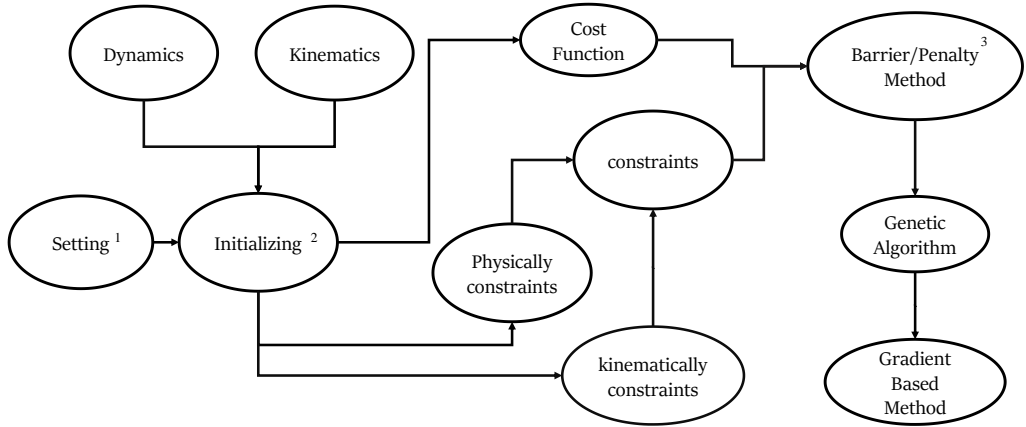

1: Setting: Type of optimization variables – Desired velocity – Initial and final configuration
2: Initialization reduces the number of variables and simplifies optimization.
3: Using penalty/barrier functions, the constrained problem becomes unconstrained.
$F(x, r) = f(x) + P(h(x), g(x), r)$
where $f(x)$ is the cost function h(x) is the vector of equalities constraint, g(x) is the vector of inequalities constraint, r is a vector of penalty parameters and $P$ is a real-valued function whose action of imposing the penalty on the cost function is controlled by r.

*Figure 3 optimization diagram*

According to equation (7), the number of unknown coefficients for a polynomial of order 4 is equal to 5. On the other hand, due to the existence of 5 independent angles, the number of unknown coefficients in the problem is 25. By determining the initial and final configuration of the robot, the number of optimization variables for this problem is reduced to 15 (by initializing).

**6. Results**

The simulation is based on the specifications of the RABBIT robot (Table 1). As a review, the nonlinear and constrained optimization problem is first converted to a non-constrained problem by the penalty method, then with the values and parameters in Tables 2 and 3, the first layer optimization problem is solved using the genetic algorithm. Next, the outputs of the first layer of optimization are considered as the start point (initial condition) of the second layer of optimization. The maximum violation of the constraints will be equal to .01 and the maximum iteration of the interior-point algorithm is equal to 20. The initial and final configuration of the system as well as other specifications and constraint bounds are given in Tables 3 and 4, respectively.

*Table 1 RABBIT parameters[18]*

| Symbol | Value | Name |
|---|---|---|
| $m_1, m_5$ | 3.2 kg | mass of lower leg |
| $m_2, m_4$ | 6.8 kg | mass of upper leg |
| $m_3$ | 20 kg | mass of trunk |
| $I_1, I_5$ | 0.93 kg-m² | rotational inertia of lower leg, about its center of mass |
| $I_2, I_4$ | 1.08 kg-m² | rotational inertia of upper leg, about its center of mass |
| $I_3$ | 2.22 kg-m² | rotational inertia of trunk, about its center of mass |
| $l_1, l_5$ | 0.4 m | length of lower leg |
| $l_2, l_4$ | 0.4 m | length of femur |
| $l_3$ | 0.625 m | length of trunk |
| $d_1, d_5$ | 0.128 m | distance from lower leg center of mass to knee |
| $d_2, d_4$ | 0.163 m | distance from upper leg center of mass to hip |
| $d_3$ | 0.2 m | distance from trunk center of mass to hip |

*Table 2 Quantities and specifications of genetic algorithms*

| Population size | 300 |
|---|---|
| Initial range | [-12,12] |
| Elite count | 15 |
| Crossover fraction | .8 |
| Migration fraction | .2 |
| Stall generation | 50 |
| Function count | 10401 |

*Table 3 Problem physical parameters and constraints*

| Maximum angular rate | 5 rad/s |
|---|---|
| Maximum actuator torque | 150 N.m |
| Step length | 0.5 m |
| Velocity | 1m/s |
| Maximum Friction coefficient | 0.7 |

*Table 4 Initial and final configuration*

| Relative angles | Initial value@(t=0) | Final value@(t=T) |
|---|---|---|
| q1 | -0.1681 | 0.4754 |
| q2 | 0.3073 | 0.3073 |
| q3 | -0.6499 | -0.0064 |
| q4 | 0.0064 | 0.6499 |
| q5 | 0.3073 | 0.3073 |

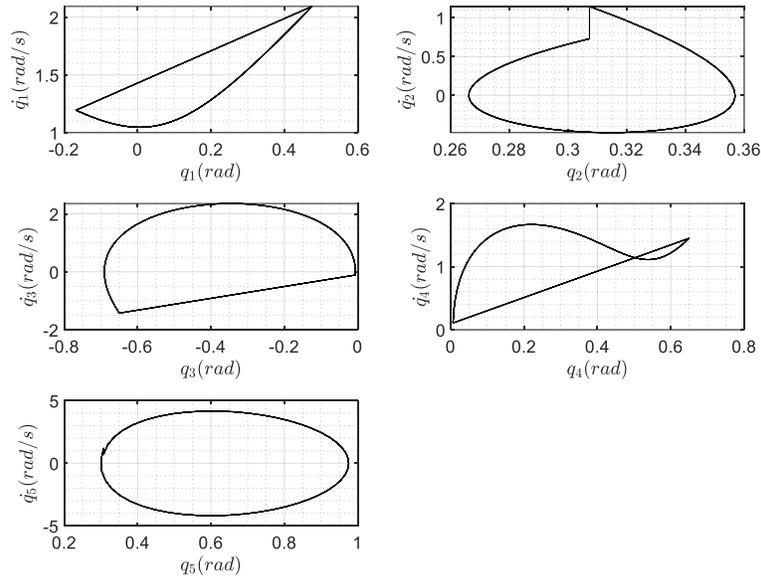

*Figure 4 The phase plots of joint angles vs. Joint angular rates*

As can be seen from the results of Figure 4, simulation results show that optimization by considering zero-dynamics constraint can produce an ideal limit cycle in walking of the biped. It is clear that angular velocities, like angles, are quite smooth and without fractures or discontinuities. They are also a long distance from their saturation limit (5 radians per second).

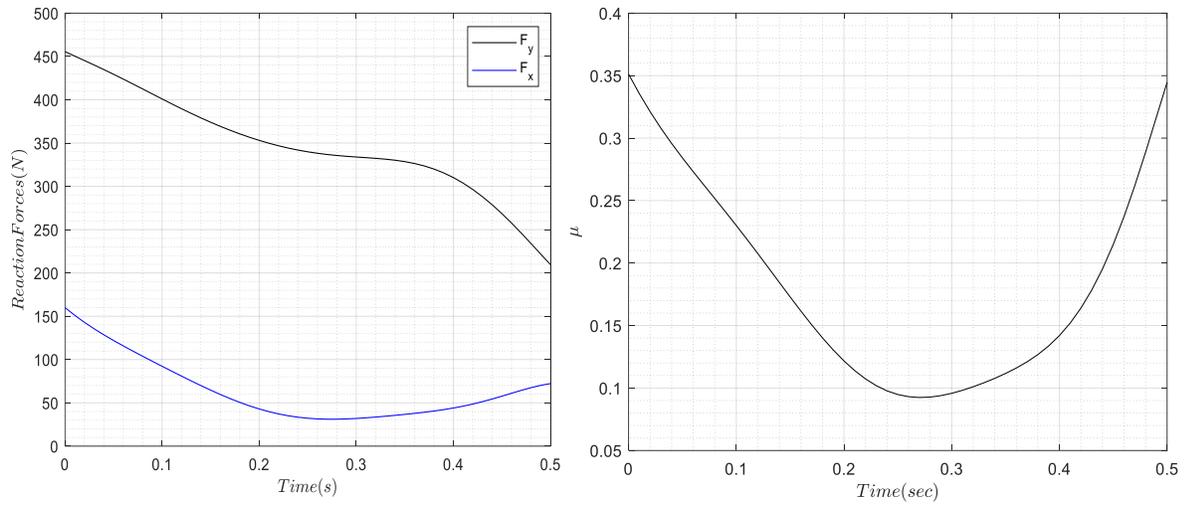

*Figure 5 Force reactions and Friction coefficient*

It is also clear from Figure 5 that the ground reaction force is also a positive value to ensure that the robot does not rise completely from the ground and the static friction coefficient required between the heels and the ground. As it is known, the coefficient of friction has desirable values that do not reach the upper bound [19].

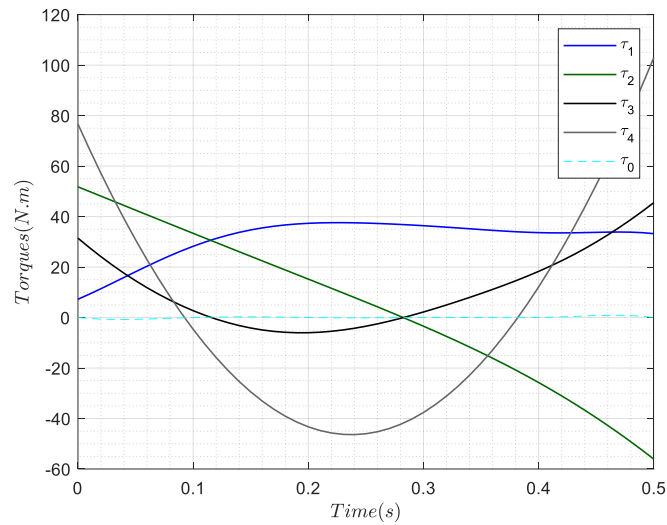

*Figure 6 Input torques*

As can be seen from figure 6, the torques are without fractures and are also far from their saturation limits.

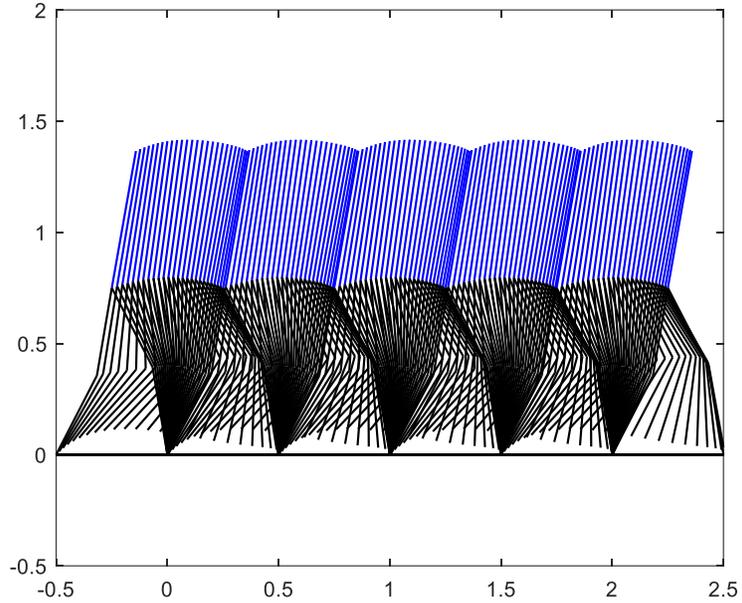

*Figure 7 Walking motion*

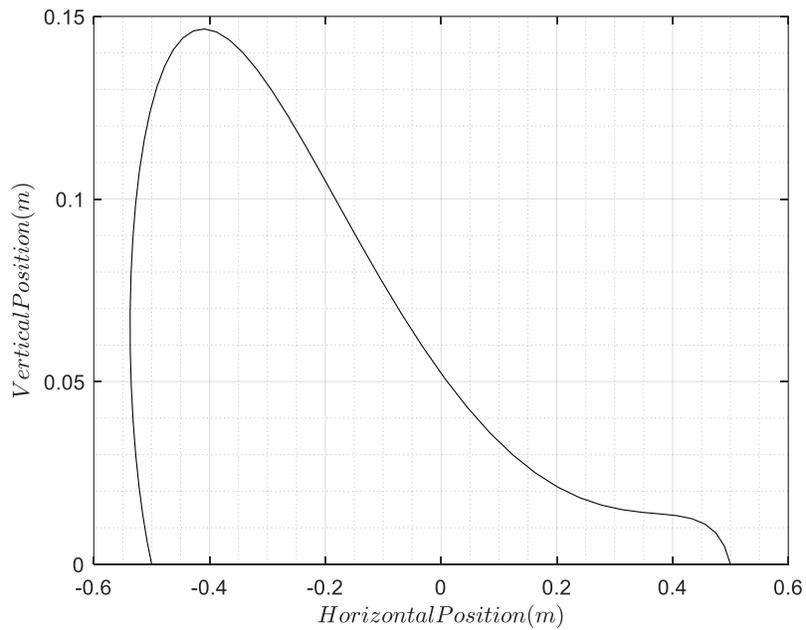

*Figure 8  Position of the swing leg's foot*

As shown in Figures 7 and 8, the swing leg does not collide with the ground except at the beginning and end of the phase.

**7. Conclusion**

This paper proposes a two-layer framework for generating optimal time-varying trajectories for bipedal robots. The novelties of the proposed work are presenting and satisfying the impact invariance constraint in a new way to ensure the periodicity of the gait in each phase and satisfying the hybrid zero dynamics simultaneously without any approximation. Also to find a better optimal solution, a hybrid optimization is used. On the other hand, various constraints were considered for a better motion of the robot. According to the simulation results, the accuracy of the proposed method and the obtained optimal solution were confirmed.